# PIKS (pruned iterative k-means searchlight): a technique to identify actionable trends for policy-makers through open healthcare data

A. Ravishankar Rao, PhD (corresponding author)

Fellow, IEEE

Fairleigh Dickinson Univ., NJ, USA

[raviraodr@gmail.com](raviraodr@gmail.com)

Subrata Garai

Independent IT Software Engineer

subratagarai@gmail.com

Soumyabrata Dey, PhD.

Clarkson University, NY, USA

[soumyabrata.dey@gmail.com](soumyabrata.dey@gmail.com)

Hang Peng

Independent Machine Learning Engineer

hanspenghang@gmail.com

*Abstract*

**With calls for increasing transparency, governments are releasing greater amounts of data in multiple domains including finance, education and healthcare. We focus on healthcare due to its economic importance worldwide. The efficient exploratory analysis of healthcare data constitutes a significant challenge. Key concerns in public health include the quick identification and analysis of trends, and the detection of outliers. This allows policies to be rapidly adapted to changing circumstances.**

**We present an efficient outlier detection technique, termed PIKS (Pruned iterative-k means searchlight), which combines an iterative k-means algorithm with a pruned searchlight based scan. We apply this technique to identify outliers in two publicly available healthcare datasets from the New York Statewide Planning and Research Cooperative System, and California's Office of Statewide Health Planning and Development. We provide a comparison of our technique with three other existing outlier detection techniques, consisting of auto-encoders, isolation forests and feature bagging.**

**We identified outliers in conditions including suicide rates, immunity disorders, social admissions, cardiomyopathies, and pregnancy in the third trimester. We demonstrate that the PIKS technique produces results consistent with other techniques such as the auto-encoder. However, the auto-encoder needs to be trained, which requires several parameters to be tuned. In comparison, the PIKS technique has far fewer parameters to tune. This makes it advantageous for fast, "out-of-the-box" data exploration. The PIKS technique is scalable and can readily ingest new datasets. Hence, it can provide valuable, up-to-date insights to citizens, patients and policy-makers.**

**We have made our code open source, and with the availability of open data, other researchers can easily reproduce and extend our work. This will help promote a deeper understanding of healthcare policies and public health issues.**

I. INTRODUCTION AND MOTIVATION

The exploratory analysis of healthcare and mortality data is an important research area, as it directly impacts the wellbeing of citizens. Even relatively simple techniques such as slicing and dicing existing data in new ways can yield remarkable insights. A very prominent recent example is the work of the Nobel Prize winner in economics, Angus Deaton, who examined mortality rates [1]. Traditionally, only two buckets were examined, consisting of people whose ages were less than 50 years, and those whose ages were greater than or equal to 50 years. Instead, Deaton partitioned the data differently into buckets of 10 years each, and specifically examined trends in the age group of 40-50 year olds. An unexpected increase in mortality rates was observed for middle-aged white Americans [1]. Some of the contributors were alcoholism and suicides. This work also shed light on the growing opioid crisis in the US which is now a leading cause of death in the younger population [2]. A noteworthy feature of this work is that it combined analytics techniques with publicly available health data to arrive at meaningful results. Hence, there is value in making healthcare data open so that these types of trends can be identified early by researchers.

A recent review paper by Jose-Sousa et. al[3] observes that a global and interconnected world has turned decision making into an increasingly complex and unpredictable endeavor. In the domain of healthcare, organizations are being challenged to operate in dynamic environments characterized by constant changes in population health, disease trends, drug availability, costs and government regulation. In this context, the use of analytics, including big-data approaches are becoming essential. Big data analytics is still a relatively new technology in the healthcare arena and less than half the number of healthcare organizations are actively applying it [4]. Though artificial intelligence techniques have been applied to medical problems for a few decades[5], healthcare organizations are approaching such techniques with caution[6]. Hence, there is room for growth in applications of artificial intelligence and big data analytics to healthcare.

A major obstacle in creating such applications is the lack of readily available data. Patient data is private and cannot be shared amongst organizations easily, even for non-commercial research purposes [7]. Furthermore, many electronic health record systems are not compatible, and considerable effort has to be expended on the pre-processing stages of data preparation [8]. Pharmaceutical companies typically do not share their data as that provides them with a closely guarded competitive advantage [9].

Since it is important to protect the privacy of patients, some states have started releasing de-identified data in open repositories, such as New York Statewide Planning and Research Cooperative System SPARCS [10] and the California Office of Statewide Health Planning and Development (OSHPD) (https://oshpd.ca.gov/). The benefits of using open datasets and open source software are that researchers can readily replicate results and build on each other's work.

There is definite value to be harvested from public datasets. Building on the work of Deaton [1], Masters et al. [11] analyzed death counts for U.S. white women and men aged 45–54 for years 1980–2013, obtained from the National Center for Health Statistic's (NCHS) Mortality Multiple Cause-of-Death Public Use Records. They demonstrated that significant gender based differences existed, which was a novel result. Knowledge of this fact can help governments tailor their services to help vulnerable segments of the population.

For future research, it is important to have open data sources and encourage researchers to publish their code and algorithms so that others can independently verify the resulting research findings. We favor this approach in our paper, and demonstrate this path by using open data sources and releasing our code to the public (see

https://github.com/fdudatamining). Our approach is consistent with the open research culture that is being proposed in other disciplines such as science[12] and psychology [13].

Given the availability of open health datasets, it is important to investigate techniques to extract value from them. The first step is to conduct an efficient exploration of these datasets, and is currently active research area [14-16]. Subsequent steps include creating predictive models through machine learning [17], and deploying these models in applications [18]. The scope of the current paper is restricted to the topic of exploratory data analysis on open health datasets.

It is essential to quickly obtain a "birds-eye" view of the data, through techniques such as visualization and clustering. This facilitates interpretation and attaching meaning to the data [19]. A key step involves the rapid identification of outliers or anomalies. Rao et al. [20] combined techniques from database analysis and machine learning to derive an iterative k-means technique to identify outliers. This was advanced by adding a subset-scanning capability [21]. The main contribution of the current paper is to expand our prior work and the state-of-the-art in three directions. Firstly, we created a more efficient algorithm, termed PIKS (pruned iterative k-means searchlight) to explore the search space of outliers through the use of pruning and feature selection. The use of pruning reduces the processing time by approximately 19%. As the number of features increases, we can expect greater reductions in the absolute processing time. Secondly, we apply the PIKS outlier detection technique to two new datasets. This consists of a newly released dataset by New York State SPARCS, and a dataset released by the California OSHPD that was not previously analyzed. We applied the PIKS technique to de-identified data from approximately 19 million patient records over a 6 year period obtained from New York SPARCS [10]. Notable outliers we detected include rises in suicides and administrative costs in New York, and cardiomyopathies and pregnancy disorders in California. This is a significant contribution, as such findings have not been previously reported from these datasets. Furthermore, these findings are consistent with results obtained through alternate methodologies utilized by other researchers. We present a detailed analysis of these and other outliers in the Results section. Thirdly, we provide a comparison of the PIKS technique with other outlier detection techniques consisting of auto-encoders, isolation forests and feature bagging.

The expanded analysis presented in this paper, and the selection of the domain of healthcare data analysis, specifically open health data, should prove valuable to researchers in the areas of exploratory data analysis, outlier detection and open data [22]. Though we tested our techniques on open health data, they are general and could be applied to structured numerical data from other domains.

II.  BACKGROUND AND RELATED WORK

Holzinger [23] observed that the areas of biomedical research, clinical practice and healthcare are becoming inundated with data, which makes the extraction of meaning and knowledge increasingly difficult. The state of the art consists of applying visualization techniques to plot the data. Popular open source packages are Seaborn, plotly and Matplotlib. A sample plot of healthcare data in the New York SPARCS dataset is shown in Figure 1. This heatmap requires aggregation over selected variables, such as hospital names and clinical diagnosis codes. Further computation such as calculating the percentage increase with respect to a baseline year may be required. The color scale needs to be chosen carefully and values may need to be thresholded to improve the dynamic range of visible information. Significant care thus needs to be utilized to provide meaningful plots. The volume of data may be too much to fit into a single viewable screen without losing important details. Finally, it is virtually impossible to manually sift through hundreds of such plots for different combinations of variables and aggregations.

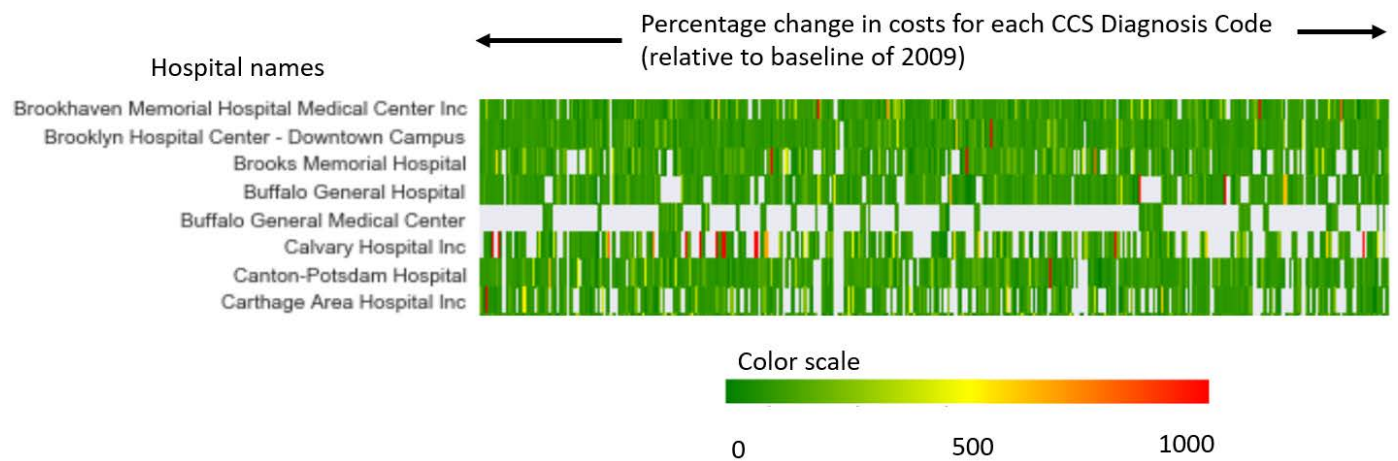

**Figure 1:** Several million patient records are aggregated by hospital names and clinical diagnosis codes (CCS codes). This visualization shows the percentage increase from the year 2009 to the year 2016 for each CCS Diagnosis code over each hospital. For instance, Calvary Hospital shows a few entries colored with red, indicating that there were high percentage increases for the corresponding CCS Diagnosis codes. We show only a few hospital names, as there are around 200 hospitals in the database.

Due to these problems noted above, there is an opportunity for the application of artificial intelligence and machine learning techniques. Exploratory analysis is a key capability that is enabled by the use of these techniques [23, 24]. It is important for the exploratory analysis to be interactive [23]. It is desirable to have a technique that will automatically sweep across the available data, aggregate it suitably, and identify interesting trends such as outliers. Hence, we focus on the use of outlier detection techniques in this paper as a concrete illustration of the use of artificial intelligence in healthcare. We demonstrate how the PIKS outlier detection can aid in the extraction of meaningful insight from large healthcare datasets.

There is a rich literature on outlier detection techniques. We will review research efforts that have applied outlier detection techniques in the field of healthcare. Seo [25] presented a review of univariate outlier detection techniques with several examples drawn from public health. One of the common issues with health data is that they tend to be skewed, for instance in measurements of blood pressure or surgical procedure times. Similarly, health expenditures have proven difficult to characterize as their distributions are usually heavily right skewed and their underlying models are nonlinear [26]. In the case of insurance claims data, some researchers have built models by thresholding the data. For instance, Cumming et al. [27] truncated claims higher than $50,000, which allowed a regression model to be built.

This may be considered one form of outlier detection or elimination. However, it requires detailed a priori knowledge of the domain being modeled, and is not generalizable to different domains. Goldstein and Uchida present an analysis of unsupervised anomaly detection techniques [28].

Auto-encoders are a popular means to address the anomaly or outlier detection problems. They operate by reconstructing the inputs with a reduced dimensional representation, and in the process, ignoring data patterns which occur sparsely. These sparsely occurring data patterns may be considered to arise from anomalies in the regular distribution of the data, and can be identified through the reconstruction error produced by the auto-encoder.

D'Avino et al. [29] used a combination of a recurrent neural network and an auto-encoder-based model to identify video forgery. Zhao et al. [30] devised a novel spatio-temporal auto-encoder for detecting anomalous events in a video sequence. Kawanishi et al. [31] used a variational auto-encoder for dementia detection as they treated the diseased data is outliers of healthy data. Karpinski et al. [32] used an auto-encoder to correct the corrupted outlier heartbeats captured from ECG signals.

The isolation forest is a tree-based approach [33], similar to decision trees, where a partition is built by randomly selecting a feature and then choosing a random split value that lies between the minimum and maximum value of the selected feature. An anomaly score is assigned to each point based on the number of partitions required in a binary search tree.

Feature bagging [34] is an ensemble-based outlier detection method that combines the scores generated by multiple individual outlier detection algorithms. A density-based approach [35] is used as the base outlier detection algorithm, and exploits the observation that points in dense neighborhoods are unlikely to be outliers.

Mihaylova [36] surveyed different statistical models to analyze healthcare data, and showed each scenario is best served with a specific model. This makes it challenging to analyze new data from emerging techniques, and creates a need for a method that is model free. Unsupervised techniques such as the PIKS technique presented in this paper constitute a viable option.

Bolton [37] used an unsupervised technique for fraud detection using break point analysis. Golmohammadi [38] presented an approach to detecting outliers in time series data by using a moving temporal window and comparing deviations with respect to the value at the centroid of this window.

Outlier detection can also be performed by using labeled data. Typically, there is an insufficient amount of labeled data available, especially in applications like fraud detection. Furthermore, supervised techniques also suffer from the problem of class imbalance, as the number of outliers tends to be very small. In the case of credit card transactions, the number of potentially fraudulent is approximately less than 1% of all transactions [39].

Krumholz [40] observed that though there are vast amounts of healthcare data that are being collected, the bottleneck lies in their analysis, which allows meaningful insights to be derived. This is true both in the realm of mining EHRs containing private patient data [41] to analyzing publicly available healthcare datasets [20, 21, 42-45]. There are multiple scenarios in these domains where it is advantageous to have a technique for quickly detecting meaningful trends and outliers.

The use of clustering techniques is quite popular in the applications of artificial intelligence to medicine, including DNA microarray analysis [46], the analysis of patient groups [47], breast cancer identification [48], gene expression analysis [49] and functional MRI [50].

The Amazon Sagemaker product on Amazon Web Services [51] uses a random-cut forest algorithm to perform anomaly detection on a dataset [52]. The algorithm is typically applied to streaming data, where a model is initially created with normal data points.

Rao and Clarke [44] applied k-means cluster analysis to determine trends in the graduation numbers for healthcare professionals in the US. The dataset consisted of records of individual practitioners associated with the Center for Medicare and Medicaid Services (CMS) in the US. The dataset is available freely to the public at https://data.medicare.gov/data/physician-compare. Using only this data and no other reports, Rao and Clarke [44] determined healthcare specialties that are enjoying growth and those that are experiencing decline. They also identified inherent clusters in the different specialties by using the k-means clustering algorithm [53]. A noteworthy trend is the significant decline of providers in psychiatry and general practice, which have policy implications. There is a predicted shortage of medical graduates in the US [54], which has led to a self-reflection in the medical community about why medical students are not pursuing careers in general practice and primary care [54].

This example shows that researchers can readily analyze the observations made by practitioners, policy makers and the popular press, provided the underlying data are made available. This is becoming increasingly important in a world where we are becoming inundated with seemingly contradictory messages [55]. We develop such an analysis capability in the current paper through the PIKS technique. We demonstrate the utility of this technique by extracting value from large open healthcare datasets. The use of such techniques helps the research community and society better understand public health issues.

## III. DESIGN

The approach in this paper extends the earlier work of Rao et al. [21]. Table 1 shows an extract from the New York SPARCS database, describing patient variables.

| | |
|---|---|
| Hospital County | Saratoga |
| Facility Name | Saratoga Hospital |
| Age Group | 18 to 29 |
| Gender | F |
| Race | White |
| Ethnicity | Not Spanish/Hispanic |
| CCS Diagnosis Description | BREAST CANCER |
| CCS Procedure Description | MASTECTOMY |
| APR Severity of Illness Description | MODERATE |
| Total Costs | $22,731.10 |

Table 1: An example of the features in SPARCS from the year 2014. A few selected features are shown, drawn from a total of 35 such features.

Figure 2 provides a schematic of a new and improved workflow in the current paper. The first stage involves data cleaning, including removing of missing values. The second stage consists of feature selection. This is domain dependent, and requires the identification of independent variables of interest. In the New York SPARCS data we have chosen, the independent variable is the cost of medical procedures. Hence, we identify features that provide information about medical costs, such as diagnosis description, severity of illnesses, race, and ethnicity. The next stage consists of the generation of a Hasse diagram [56] for the power-set selected features as shown in Figure 3. Subsequently, a split-apply-combine operation is applied to each element of the power-set to generate aggregated measures. For instance, a measure could be the average cost of a medical procedure at each hospital. An alternate measure could be the average cost of a medical procedure organized by race and ethnicity. The iterative k-means algorithm is applied to each of these measures to detect outliers. In general, the iterative k-means algorithm can be applied to each element of the power-set of selected features. In order to reduce computation time, we propose a heuristic that consists of a pruning step. By observing Figure 3, we note that if no outliers are found at a given node, it is also likely that no outliers will be found at connected nodes at subsequent levels of the Hasse diagram [56]. This heuristic is supported through empirical observations. Since the goal of our work is an exploration of outliers, it is advantageous to focus on outliers that can be quickly identified. Given more computational resources, we can perform more exhaustive exploration at a later stage.

This entire process can be viewed as a searchlight, borrowing a term in the existing literature [57]. Since we introduce a pruning step, the composite algorithm is termed the Pruned Iterative k-Means Searchlight, or PIKS.

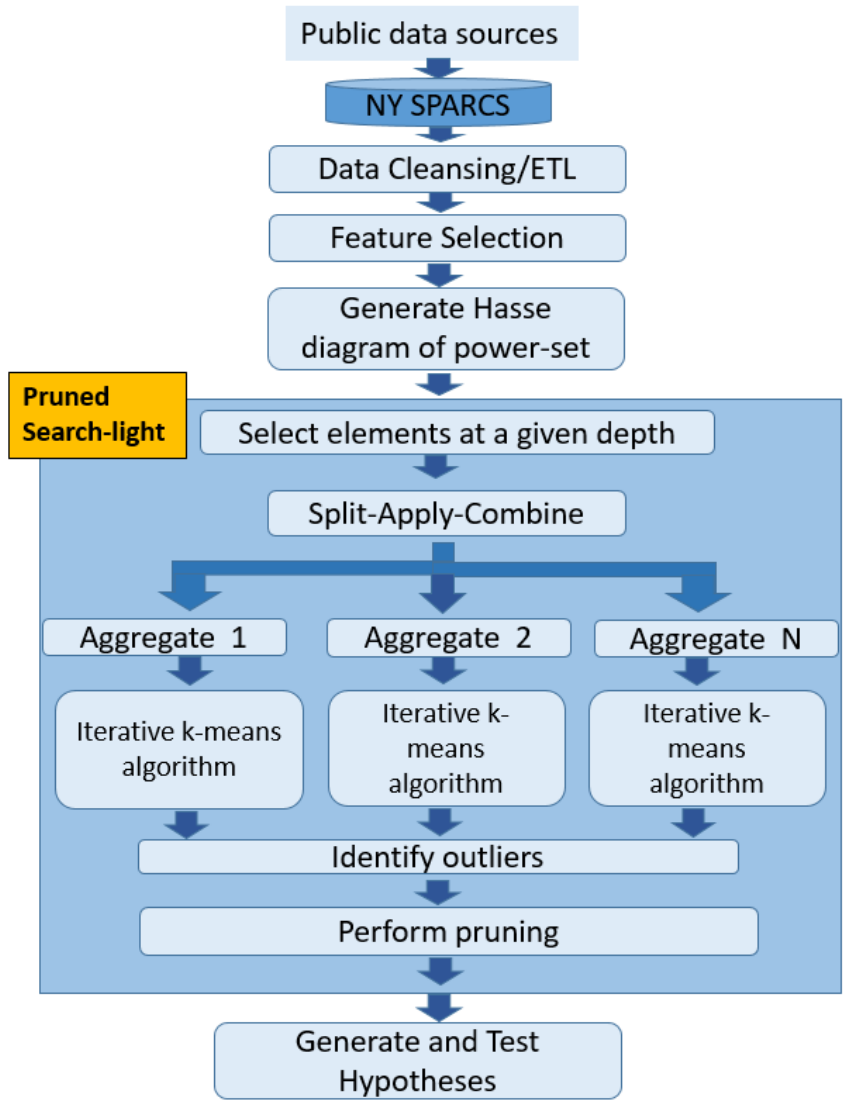

Figure 2: Proposed workflow for data analysis that produces outliers.

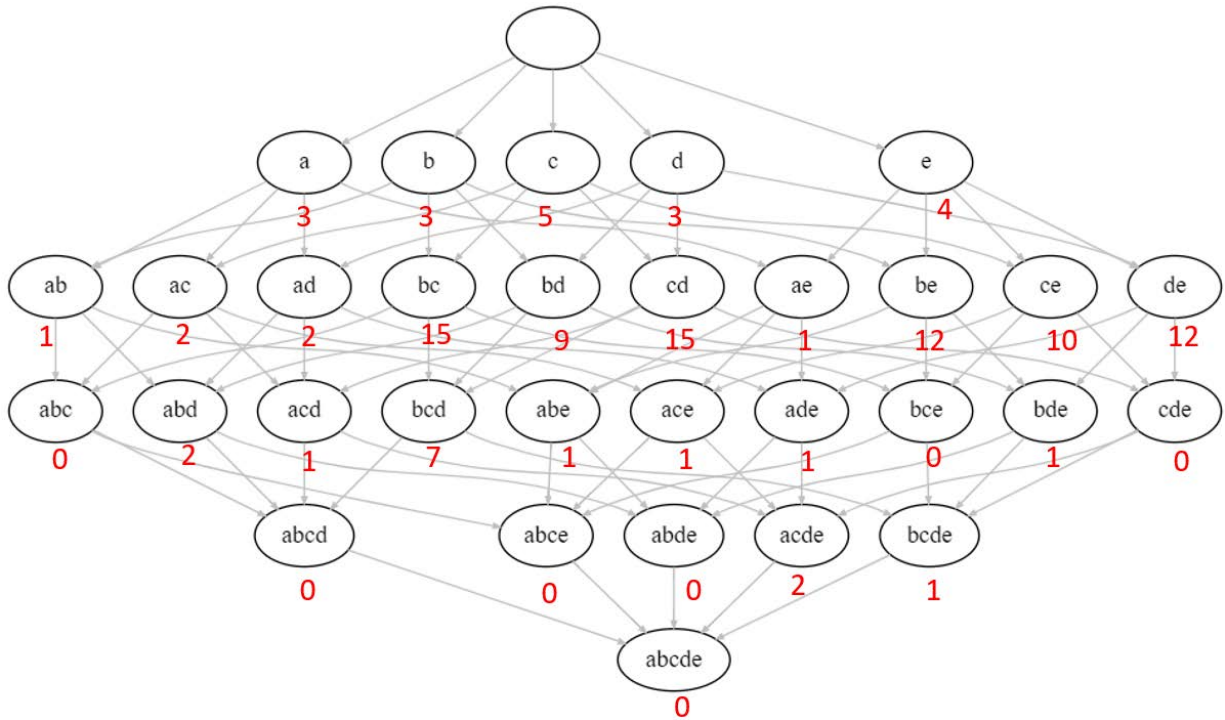

**Figure 3:** A Hasse diagram representing the relationship between elements of the power-set of S = [58 d, e]. The letters a,b,c,d,e stand for the following features. a: 'CCS Diagnosis Description', b: 'Race', c: 'Age Group' , d: 'Ethnicity', e: 'APR Severity of Illness Description'. The numbers in red represent the number of outliers found through the iterative k-means algorithm.

Several interesting use cases can be explored through a fast and robust outlier detection technique such as PIKS. For instance, outliers in disease trends can signal areas that policy makers need to focus on. Individuals can use this information when considering medical treatment or making personal health choices.

Our implementation uses the following open source components: Python Pandas, Scikit-Learn [59] and Matplotlib [60]. The Scikit-Learn Python library provides several machine-learning capabilities such as clustering, classification and prediction.

The SPARCS dataset used in our analysis consists of de-identified in-patient discharge information about disease diagnoses, costs and locations where care was provided [10]. The entire dataset from 2009-2016 [61] contains 17,559,432 rows of patient data from 254 hospitals. There are 40 columns per row, containing descriptions as shown in Table 1.

The earlier work of Rao et al. [20, 21] used SPARCS data from 2009-2014. Subsequently, newer datasets for 2015 and 2016 were released, which are analyzed in the current paper.

Table 1 shows some of the features in the original dataset, such as hospital county, Clinical Classification Software (CCS) Diagnosis Description, or Facility Name for each entry. We perform an aggregation over each of these features, as shown in Figure 2.

In this paper, we analyzed a second dataset, consisting of hospital discharge data from the California Office of Statewide Health Planning and Development (OSHPD). The data is publicly available at https://oshpd.ca.gov/, and samples are shown in Table 2 and Table 3. One of the challenges in using this data is that different ICD codes have been deployed. Hence, we mapped ICD-9 codes to ICD-10 codes in order to maintain consistency among entries in the data. This is a data cleaning and normalization step. This allows us to determine trends in the data that can be observed over multiple years.

| ICDCMCode | DiagnosisDesc | TotalDiag |
|---|---|---|
| 331.0 | ALZHEIMER'S DISEASE | 1 |
| A02.0 | SALMONELLA ENTERITIS | 616 |
| A02.1 | SALMONELLA SEPSIS | 229 |

Table 2: The data from the California Office of Statewide Health Planning and Development (OSHPD) consists of ICD codes (ICDCMCode) along with the number of diagnoses (TotalDiag) for each code per year. The data shown here is an extract from the year 2017, where ICD-10 codes were used.

| ICD9CMCode | TotalDiag |
|---|---|
| 001.9 | 1 |
| 003.0 | 493 |
| 003.1 | 182 |

Table 3: An extract from the California OSHPD data for the year 2012, where ICD-9 codes were used.

The counts for the Year 2012 were considered to be the baseline, and the counts for all the other years data were converted to a relative increase or decrease with respect to this baseline, expressed as a percentage. This results in a scaling or normalization of the data, and facilitates comparisons.

## IV. METHODS

Rao et al. used a subset-scan technique [21] to detect outliers, which was based on the iterative k-means algorithm[20]. We have improved upon this previous research by adding two processing stages consisting of feature selection and pruned subset scanning, as shown in Figure 2.

In the current paper, we apply the PIKS technique to an updated New York SPARCS dataset that includes more recent years that were not included in [21]. In addition, we apply this technique to a completely new dataset provided by the California OSHPD. Finally, we compare the performance of the PIKS technique with an auto-encoder configured to detect outliers.

### A. Feature Selection

We identified the "Total Costs" in Table 1 as an independent variable of interest in the New York SPARCS dataset. In general, the choice of independent variable is a function of a chosen dataset and knowledge of the specific application domain. We used the chi squared and mutual information measures of feature importance to identify the following features of interest: CCS Diagnosis Description (which is equivalent to CCS Diagnosis Code), age group, race, and ethnicity. Once the reader understands our methodology by using these features, it can be applied to other chosen features in this or other domains. Further details about the feature selection technique can be found in [62].

### B. Subset-scan technique.

We used a variation of subset-scan technique presented in [21]. The basic idea is to iteratively run the k-means clustering technique as shown in Figure 4. During each iteration, we treat small clusters of points as outliers and remove them. We repeat this process until no small clusters are present, or until we reach a fixed number of iterations. The remaining clusters capture the most relevant groups in the original data after outlier removal. Figure 4 uses hypothetical data to illustrate this procedure. Four iterations are shown, labeled as Step (a) - (d). We first choose the number of clusters, e.g. k=4 and perform the following steps.

1. Apply the k-means clustering algorithm.
2. If single-element or very small clusters (e.g. size <= 2) exist, treat these as outliers and remove them from the dataset. Continue processing the remaining data.
3. If no small clusters remain, terminate the iterative k-means algorithm,

The k-means algorithm is sensitive towards outliers, which are then identified by our algorithm to belong to an outlier set. We require that the number of clusters be specified, but there is no limit on the number of outliers that can exist at each iteration. We execute our algorithm for multiple iterations under user control. Sample code was presented in Rao [20].

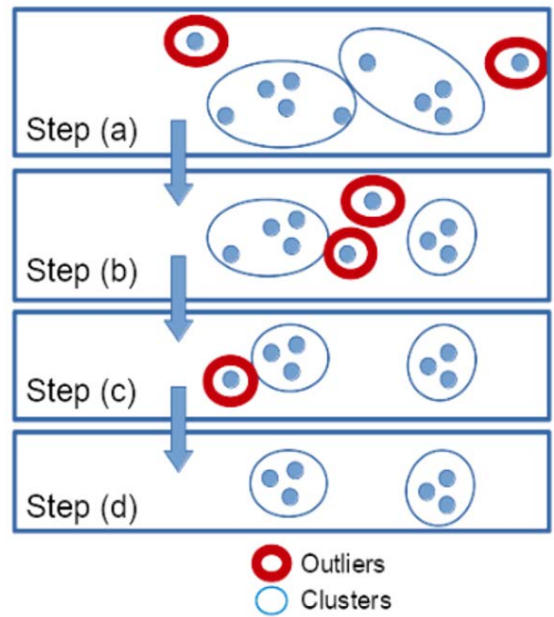

Figure 4: Shows the operation of the iterative k-means algorithm works via steps (a)-(d). Individual data points are shown as blue dots. The red circles denote outliers that are not considered for further analysis. The blue ellipses identify the remaining clusters which are processed in the next iteration.

The workflow in Figure 1 uses the iterative k-means clustering algorithm, as described above. Each field of the data, e.g. age group, gender, and race in Table 1 is referred to as a feature or a variable. A subset of the feature space is created by selecting specific features for analysis. The searchlight algorithm performs a hierarchical search over the subset of features to identify outliers. As shown in Figure 3, at the first level of hierarchical search, we use all possible feature subsets of cardinality one. At each subsequent level, subset cardinality is increased by one. As a result, each node of the hierarchical search tree is assigned to a unique feature subset. At each node, we perform a grouping of the rows of the dataset for the assigned feature subset and produce aggregate values. For example, at the first node of level 2 (Figure 3), we count the occurrences of each distinct CCS Diagnosis Description and Race tuples for the SPARCS data. The aggregated counts are further binned over the field "Discharge Year". This produces vectors of aggregate values that are scaled to find the percentage of changes with respect to the base year (2009 in the case of the SPARCS data). We will refer to vectors as 'feature vectors' for the rest of the paper. We apply the iterative k-means clustering on the feature vectors to find outliers. For instance, an outlier can be found for CCS Diagnosis Description with the label 'Suicide' and Race 'White'. A similar procedure is used for the California OSHPD data.

Searching for outliers at each node of the Hasse diagram can be computationally expensive. We propose a novel pruning step that can improve the processing speed. If at a node of the Hasse diagram no outlier is found after the iterative k-means clustering, the algorithm doesn't traverse its child node. The algorithm is summarized as follows.

## C: Pruned Iterative-kmeans Searchlight (PIKS) Algorithm

Given a set S of N features, the power-set of S contains $2^N$ elements. Each element in the power-set represents a unique feature subset as shown in the Hasse diagram in Figure 3. In discrete mathematics, the Hasse diagram is a well-known graphical representation of the relationship between elements of a partially ordered set. In this representation, the subsets at a given depth in the graph have the same number of elements. There is an implied vertical ordering, where a child node is derived from its parent nodes. For instance, the subset {a,b} can be considered to be a child node derived from the parent subsets {a} and {b}, and this is indicated by arrows between these subsets.

We traverse the nodes of the Hasse diagram in a breadth first fashion and apply the iterative k-means algorithm to the features at a given node. If a given node has zero outliers, we skip all child nodes in the Hasse diagram and do not process outliers at that node.

Each node in the Hasse diagram is associated with a Boolean label **Done.** If **Done** is 0, it means the node is waiting to be processed. If **Done** is 1 it means the node has either been processed, or has been identified as not requiring processing. All nodes are initialized with the value **Done** = 0.

Let **Num** be a binary number in the range $[0, 2^N - 1]$ representing one of the subsets in the powerset. We define a function "Execute" which takes **Num** as an argument and computes the number of outliers using the iterative k-means clustering algorithm.

Execute(**Num**):
1. If **Done** is 1, skip the processing of this node. Else proceed to step 2.
2. **S'** is the subset corresponding to **Num**
3. Use the split-apply-combine paradigm [27] to group the data by **S'**
4. Produce the aggregate values for each set of the grouped items and binned over 'Discharge Year' to produce feature vectors.
5. Apply iterative k-means algorithm on the feature vectors to find outliers.
6. Store outliers and outliers count
7. If the outlier count is 0, mark the state of all subsequent child nodes to **Done = 1**.
8. Mark the state of the current node to **Done = 1**.

The PIKS algorithm can be summarized as follows.

1. Traverse the Hasse diagram in a breadth-first fashion.
2. For each node that is visited, call Execute(Num).

3. Stop when the entire Hasse diagram is processed.

The Pandas pivot table functionality is used to execute step 3 of Execute(**Num**). The output from step 3 of PIKS can be presented visually to the user, facilitating the exploration of unusual trends, as described in Section V.

As shown in Figure 3, we use a heuristic to prune the search space. Since the node "abc" produced zero outliers, we can skip the processing of subsequent child nodes "abcd" and "abce", or prune this branch. This speeds up the processing.

We compare our technique with other approaches for outlier detection in the literature as follows.

**D. Implementation of an Auto-encoder**. An auto-encoder is an artificial neural network architecture designed for unsupervised learning from data. Auto-encoders are often used as a tool to detect outliers. An auto-encoder tries to predict its input, and in that process, first compresses the data, and later decompresses it to reconstruct the original data.

We used the H2O package (http://h2o.ai/resources/) which contains an auto-encoder function. We prepared each of the two datasets separately, consisting of the New York SPARCS data and the California OSHPD data. For the SPARCS data, we passed the dataset aggregated over the feature "CCS Diagnosis Description" to the auto-encoder estimator function (H2OAutoencoderEstimator) with following parameters: activation function as "Tanh", number of hidden layers as 10 with size epochs as 50. Alternately, a MATLAB implementation of an auto-encoder is available via the function "trainAutoencoder()".

We investigated different numbers of hidden layers and epochs in order to determine the optimum number. We observed that an auto-encoder is powerful enough to distinguish the outliers even with very low numbers of hidden layers and epochs. We selected a 10-layered network whose size is comparable to the 8-layered network used in earlier work by Rao and Clarke [43] for prediction purposes. We note that this selection is part of the parameter tuning process that accompanies the use of auto-encoders.

C. **Implementation of isolation forests and feature bagging**

We used the package PyOD (Python Outlier Detection) [63] which contains the isolation forest and feature bagging techniques for outlier detection. The basic working of these techniques was described earlier in the background section.

## V. RESULTS

We investigated multiple outlier detection techniques, including the PIKS technique, the auto-encoder, isolation forests and feature bagging. We used two different datasets, described in the sub-sections below. We set k=8 for the iterative k-means algorithm utilized within the searchlight subset scan. The results are organized according to the datasets.

A. *Analyzing the New York SPARCS dataset with the PIKS technique*

Though the PIKS technique works on an arbitrary number of dimensions, we present a pictorial representation of its application to four dimensions depicted in Figure 5. We note that this type of representation derives from prior work on association rule mining in the field of data mining [64] and is similar to the Hasse diagram in Figure 3.

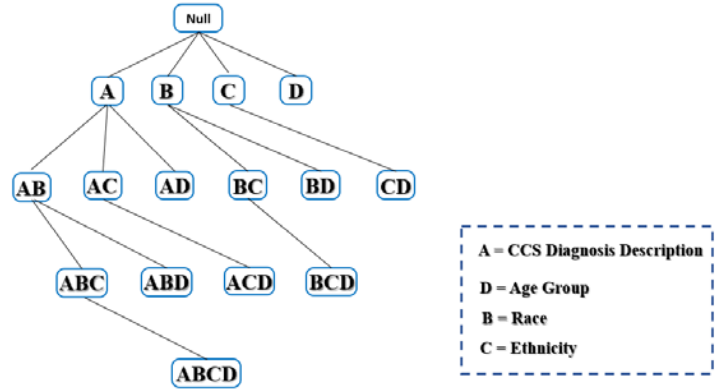

Figure 5: This shows the operation of the PIKS technique on four dimensions from the New York SPARCS dataset. The dimensions of "CCS Diagnosis Description", "Age Group", "Race" and "Ethnicity" are combined and scanned in the order shown in the above graph.

The application of the PIKS technique on the dimensions shown in Figure 5 can be further characterized by the results produced on the New York SPARCS dataset. This is shown in Figure 6, where highlighting is used to represent different results. For instance, at the second aggregation level for the label A, "CCS Diagnosis Description", isolated outliers are found in nodes with labels AC and AD only and not in node AB. This depiction provides an overall summary of the outlier results.

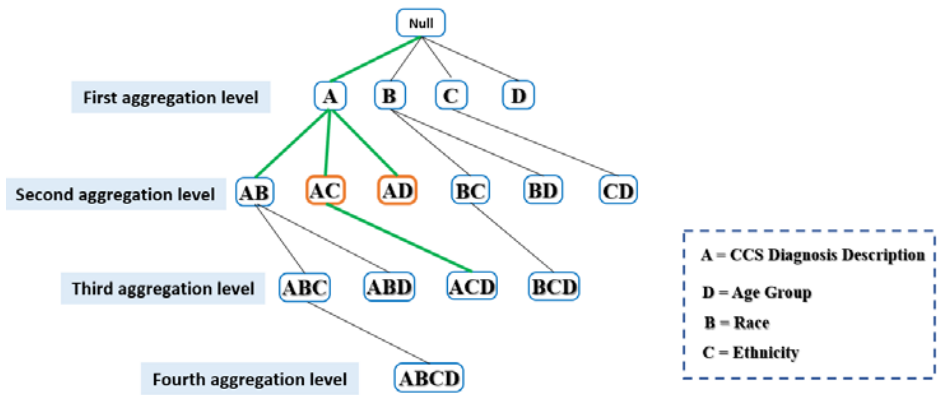

Figure 6: The aggregation levels are numbered at the left of the figure, and represent an increasing level of detail. The edge highlights indicate that the same outlier is found at different aggregate levels. The node highlight indicate that an isolated outlier is found at that level only.

Further detail is provided in Table 4. We are performing aggregations over the number of individual hospitalization incidents into different bins or nodes shown in Figure 5 and Figure 6. The column entitled "Row Counts" represents the number of hospitalizations, which can vary depending on the granularity of the aggregation performed. If the row count is less than a threshold, which was 50, we do not apply the iterative k-means algorithm on that aggregation level.

| # | Aggregation Levels | Variable Used on 'Count of Hospitalization Incident' against 'Discharge Year' | Row Counts | Qualified for Iterative k-means | Outliers Found |
|---|---|---|---|---|---|
| 1 | First Level | Diagnosis Description | 262 | Yes | INFLUENZA ADMIN/SOCIAL ADMISSION |
| 2 | First Level | Race | 3 | No | |
| 3 | First Level | Ethnicity | 3 | No | |
| 4 | First Level | Age Group | 5 | No | |
| 5 | Second Level | Diagnosis Description + Race | 785 | Yes | ADMIN/SOCIAL ADMISSION | White |
| 6 | Second Level | Diagnosis Description + Ethnicity | 780 | Yes | SHOCK | Unknown ADMIN/SOCIAL ADMISSION | Spanish/Hispanic |
| 7 | Second Level | Diagnosis Description + Age Group | 1203 | Yes | GANGRENE | 18 to 29 ADMIN/SOCIAL ADMISSION | 70 or Older |
| 8 | Second Level | Race + Ethnicity | 9 | No | |
| 9 | Second Level | Race + Age Group | 15 | No | |
| 10 | Second Level | Ethnicity + Age Group | 15 | No | |
| 11 | Third Level | Diagnosis Description + Race + Ethnicity | 2181 | Yes | MAL-POSITION/PRESNTATN | Unknown | Black/African American |
| 12 | Third Level | Diagnosis Description + Race + Age Group | 3449 | Yes | |
| 13 | Third Level | Diagnosis Description + Ethnicity + Age Group | 3261 | Yes | ADMIN/SOCIAL ADMISSION | Not Span/Hispanic | 70 or Older |
| 14 | Third Level | Race + Ethnicity + Age Group | 45 | No | |
| 15 | Fourth Level | Diagnosis Description + Race + Ethnicity + Age Group | 8071 | Yes | |

Table 4: This table describes the process used to arrive at the aggregation counts at different levels of the tree depicted in Figure 5 and Figure 6. The iterative k-means algorithm is applied only if an aggregation count exceeds a threshold, which is 50.

Next, we provide a visual depiction of the outliers, as shown in Figure 7. The percentage change of counts with respect to a baseline year is displayed on the y-axis, and the years are displayed on the x-axis. The outliers are identified as "Influenza" and "Administrative/Social Admission". Note that both positive and negative increases can be flagged as outliers. A detailed interpretation of the results is provided in the Discussion section.

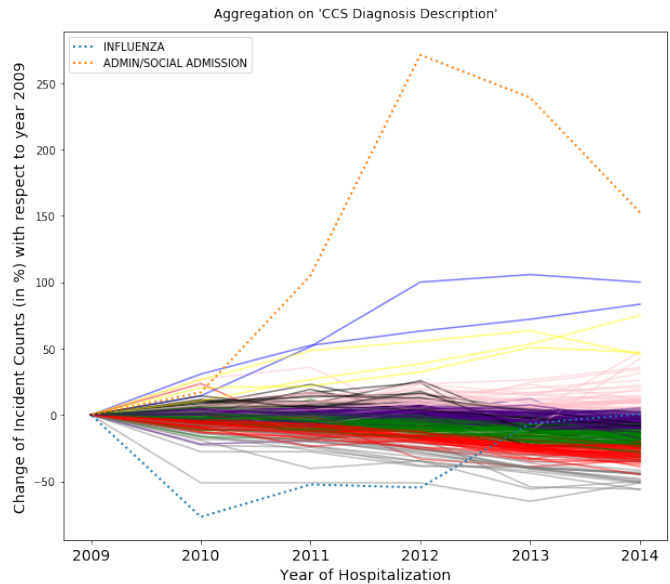

Figure 7: This figure shows the result of applying the iterative k-means algorithm during the first aggregation level of the PIKS technique. The different clusters are color coded, and the outliers are represented by dotted lines.

In Figure 8, we show the second aggregation level as described in Figure 6, using race as the second level.

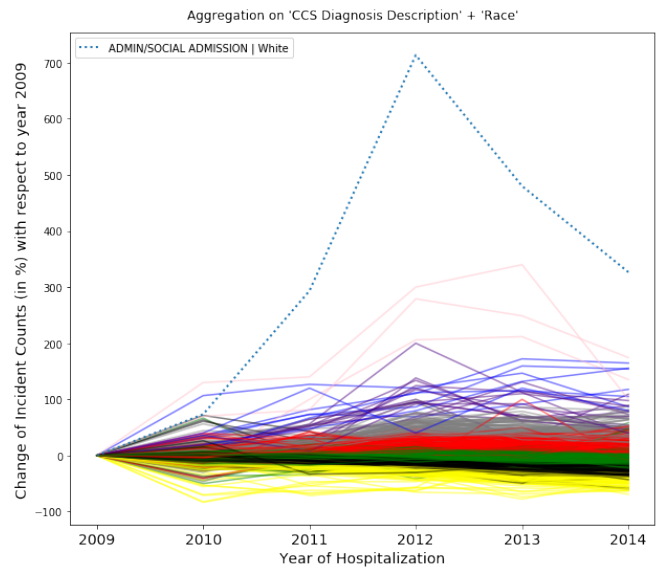

Figure 8: This figure shows the aggregation of CCS diagnoses codes along the feature of "Race" in the New York SPARCS database. The outlier is identified in the dotted blue line. The different clusters are color coded.

In Figure 9, we show the second aggregation level as described in Figure 6, using ethnicity as the second level.

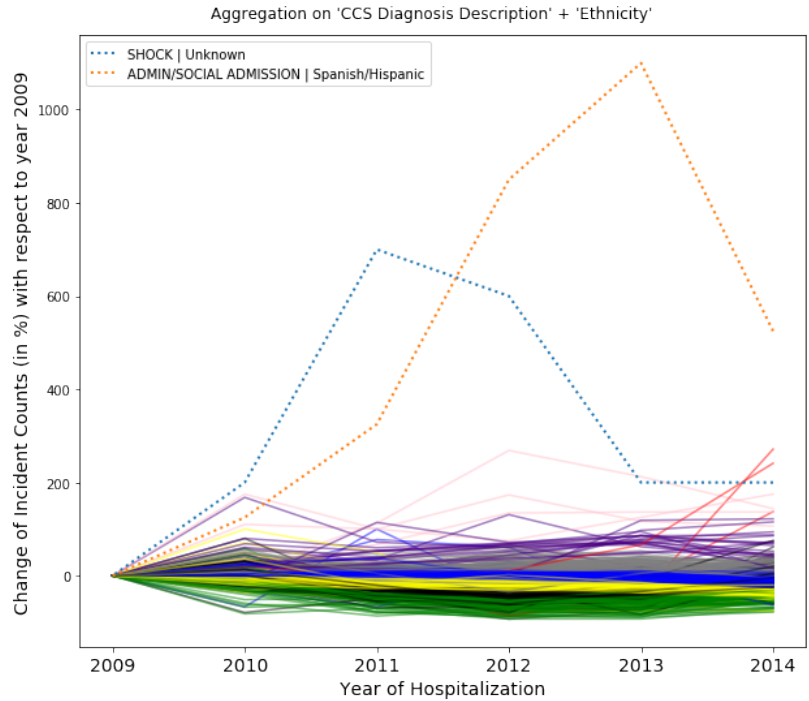

Figure 9: This figure shows the aggregation of the CCS diagnosis codes for the feature of "Ethnicity". Outliers are shown in dotted lines.

In Figure 10, we show the second aggregation level as described in Figure 6.

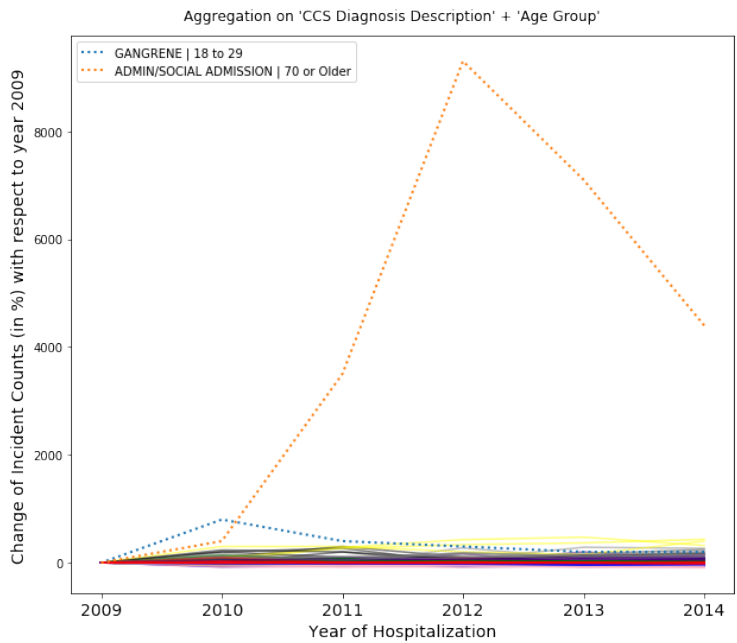

Figure 10: This figure shows the aggregation of the CCS diagnosis codes for the feature of "Age Group". Outliers are shown in dotted lines.

In Figure 11, we show the third aggregation level as described in Figure 6.

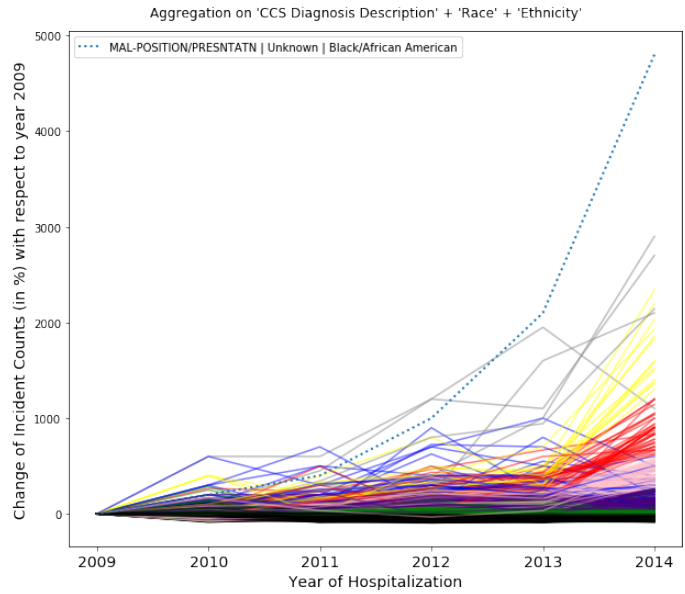

Figure 11: This figure shows the aggregation of the CCS diagnosis codes for the feature of "Race" followed by "Ethnicity". Outliers are shown in dotted lines.

In Figure 12, we show the third aggregation level as described in Figure 6, with different features.

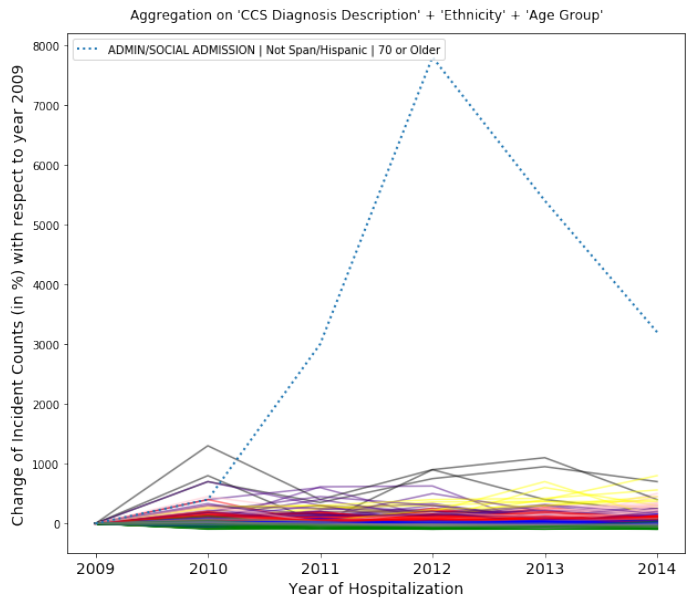

Figure 12: This figure shows the aggregation of the CCS diagnosis codes for the feature of "Ethnicity" followed by "Age Group". Outliers are shown in dotted lines.

Our analysis can be conveniently re-run for new data released by SPARCS on a continuous basis. Our earlier results cover the years of available SPARCS data at that time from 2009-2014. We now show results for the SPARCS data from the years 2009-2016. The purpose of this is to demonstrate that the outliers can change from year to year, and policy makers need to continuously update the allocation of resources based on new data.

For the years 2009-2016, we show the results for a single branch of the search tree, which starts with the grouping variable 'CCS Diagnosis Descriptions'. We create feature vectors by computing the percentage of changes in total counts of incidences for medical procedures reported in New York State during the 2009-2016 time period, where 2009 is the baseline. The clusters and outliers are plotted in Figure 13.

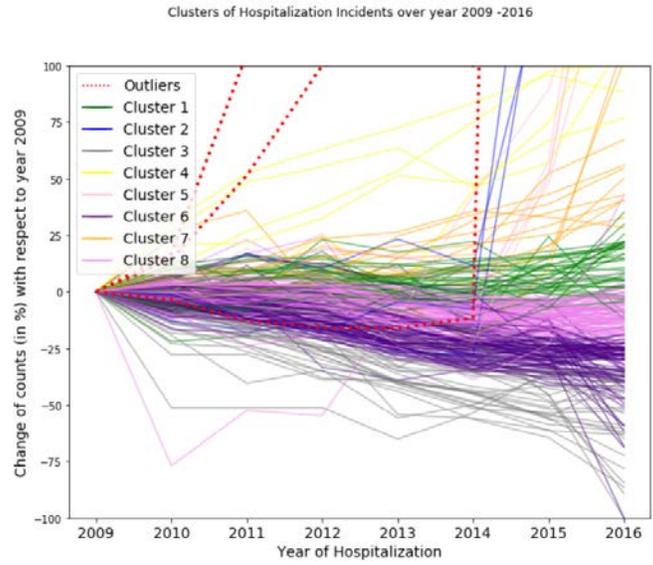

Figure 13: The figure shows the outputs of the PIKS technique in the form of clusters when the feature vectors are formed using 'CCS Diagnosis Descriptions' as the grouping variable. Each broken line in the plot represents a single CCS Diagnosis description. The outliers are shown in dotted red lines, and their descriptions are given in Table 1.

Table 5 shows the four outliers found in the first iteration of the PIKS technique depicted in Figure 13.

| Outliers detected through the PIKS technique |
| --- |
| Administrative/social admission |
| Immunity disorders |
| Suicide and intentional self-inflicted injury |
| Influenza |

Table 5: This table shows the labels of the outliers found through the application of PIKS technique on the New York SPARCS data.

Note that this is just one view of the many outliers found in the New York SPARCS dataset. Other views of outliers are presented as follows. In the next series of plots, we compute the average cost of diagnosis for each CCS Classification Code across all hospitals in the New York SPARCS dataset for the year 2015. The feature vector for a hospital consists of the average cost of diagnosis for each CCS Classification Code. We then apply the PIKS technique to these feature vectors to identify outliers. The results are presented in Figure 14(A) and (B).

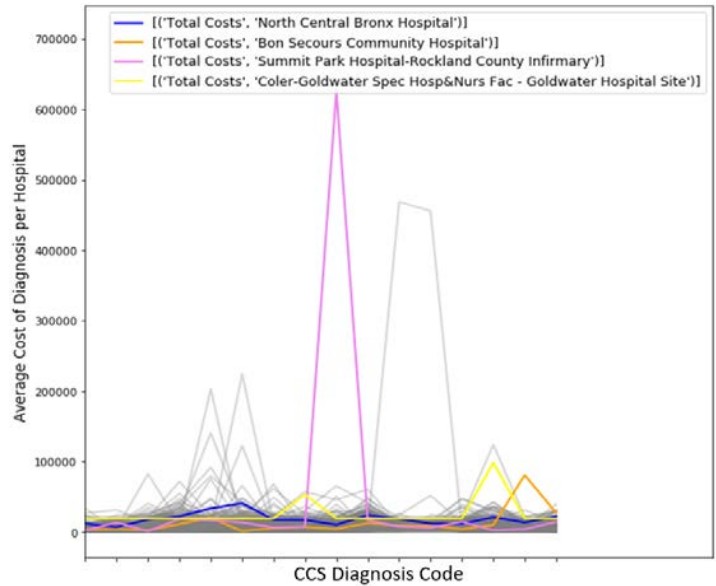

Figure 14: (A) This figure shows the distribution of average costs of each procedure (CCS Diagnosis Description) across all the hospitals. Each distribution is color coded according to its cluster number assigned by the PIKS technique. Since it is difficult to show all the 255 CCS Diagnosis codes in a single plot, we have used the plot in (B) to depict the additional CCS Diagnosis codes. The top four outliers are color coded according to the legend in the rectangular box.

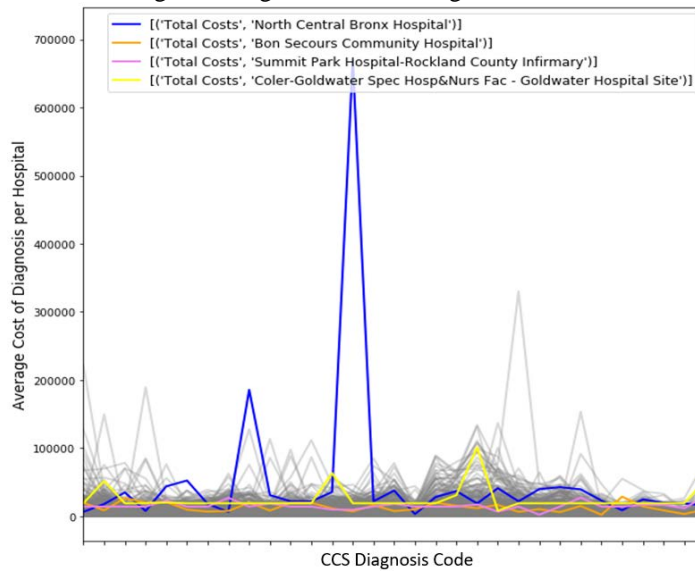

Figure 14 (B) This figure is continued from Figure 14 (A) and shows the distribution of average costs of each procedure (CCS Diagnosis Description) across all the hospitals. Each distribution is color coded according to its cluster number assigned by the PIKS technique. The top four outliers are color coded according to the legend in the rectangular box, and are consistent with the coloring in Figure 14 (A).

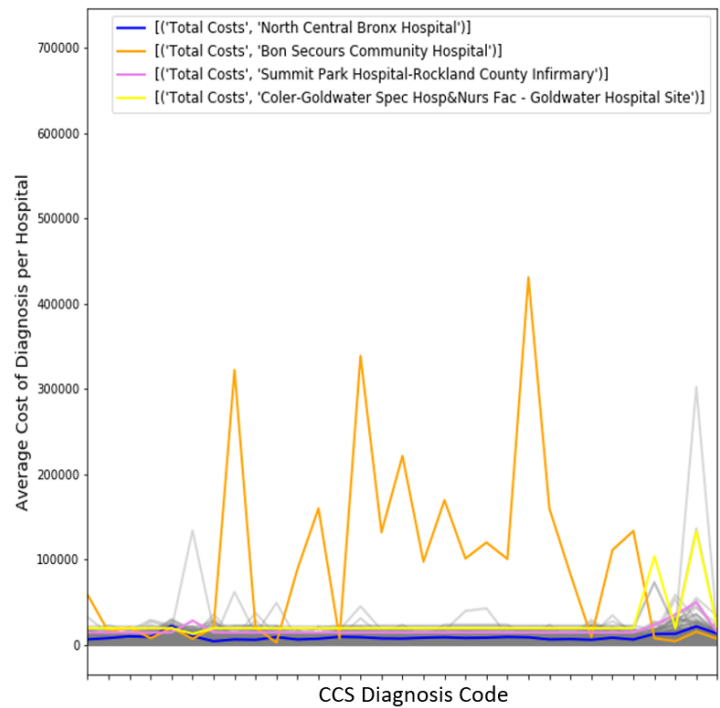

Figure 14 (C) This figure is continued from Figure 14 (B) and shows the distribution of average costs of each procedure (CCS Diagnosis Description) across all the hospitals. Each distribution is color coded according to its cluster number assigned by the PIKS technique. The top four outliers are color coded according to the legend in the rectangular box, and are consistent with the coloring in Figure 14 (A).

Other combinations of variables in the aggregation step shown in Figure 2 can be used to detect different types of outliers. For instance, an aggregation of costs across counties, revealed that Westchester county has the largest cost increase for mental health diseases [45].

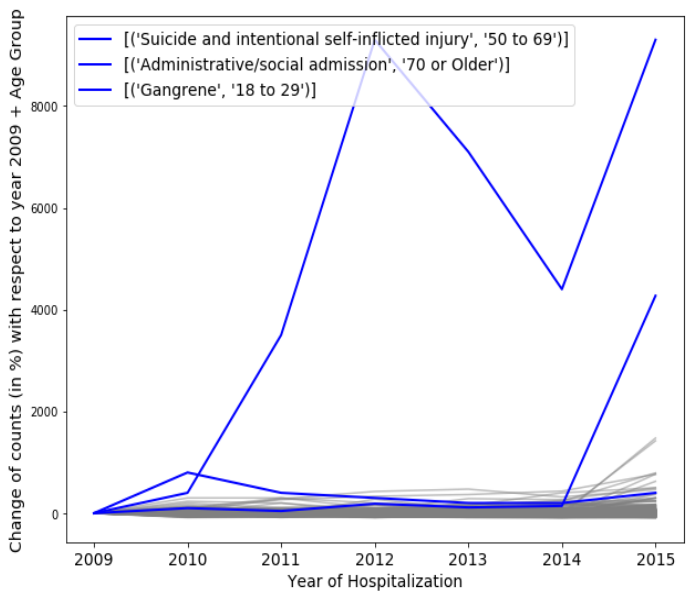

Figure 15: Shows the outliers in the CCS Diagnosis Codes when they are segmented by Age Groups. One of the outliers consists of Suicides in the age group 50-69 years.

As shown in Figure 3, and in the implementation section IV-C, the pruning step produces a reduction in the computation time of approximately 20%.

B. *Analyzing the New York SPARCS dataset using auto-encoders*

In Table 6 we present a comparison of the outliers detected for the New York SPARCS data from the year 2009-2014, as shown in the previous figures.

| Outliers detected through the searchlight subset scan technique (SPARCS data from 2009-2014) | Outliers detected by the auto-encoder (SPARCS data from 2009-2014) |
|---|---|
| Administrative/social admission | Administrative/social admission |
| Administrative/social admission\|70 or older | Acquired Foot Deformity \| Unknown \| Other Race |
| Administrative/social admission\|Not Span/Hispanic\|70 or older | Acute Posthmrg Anemia |
| Administrative/social admission\|White | Administrative/social admission\|70 or older |
| Administrative/social admission\|White | Administrative/social admission\|Not Span/Hispanic\|70 or older |
| Gangrene\|18 to 29 | Administrative/social admission\|Spanish/Hispanic |
| Immunity disorders | Administrative/social admission\|White |
| Influenza | Gangrene \| 18 to 29 |
| Mal-Position/Presentation\|Unknown\|Black, African American | Hyperlipidemia \| 30 to 49 |
| Shock\|Unknown | Immunity Disorder\|Not Span/Hispanic |
|  | Immunity Disorder\|Other Race |
|  | Influenza |
|  | Mal-Position/Presentation\|Unknown\|Black, African American |
|  | Medical Exam/Evaluation \| 70 or Older |
|  | Medical Exam/Evaluation \| Spanish/Hispanic \| 18 to 29 |
|  | Other Urinary Cancer \| Spanish/Hispanic \| White |
|  | Shock\|Unknown |

Table 6: The outliers detected through the PIKS technique are compared with the outliers detected through the auto-encoder. In ech column, the entries are sorted alphabetically.

In Table 7 we show that the outliers detected by the PIKS technique are consistent with the outliers detected by the auto-encoder against the SPARCS data from 2009-2015. Only the first level aggregations are shown for the sake of brevity.

| Outliers detected through the searchlight subset scan technique (SPARCS data from 2009-2014) | Outliers detected by the auto-encoder (SPARCS data from 2009-2014) |
|---|---|
| Administrative/social admission | Administrative/social admission |
| Immunity disorders | Suicide and intentional self-inflicted injury |
| Suicide and intentional self-inflicted injury | Immunity disorders |
| Influenza | |

Table 7: The New York SPARCS dataset was analyzed. We compare the labels of outliers detected after the first iteration of PIKS technique (shown earlier in Table 5) with the outliers detected by the auto-encoder. The SPARCS data cover the years 2009-2015.

We present a visualization of the operation of the auto-encoder used for SPARCS data from 2009-2015 in Figure 16.

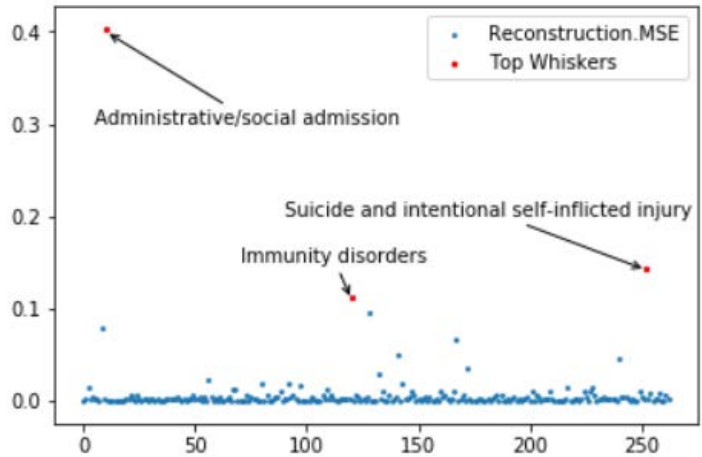

Figure 16: Shows the result of applying the auto-encoder to the New York SPARCS data. Outliers are shown in red dots along with the disease descriptions. The x-axis contains the feature index, and the y-axis depicts the reconstruction error.

### C. *Using the isolation forest and feature bagging outlier detection techniques*

In Table 8, we compare the application of the isolation forest and feature bagging outlier detection techniques on the New York SPARCS data from 2009-2014. We used the PyOD outlier detection package to perform the computations.

|     | Isolation Forest | | | Feature Bagging | | |
| --- | --- | --- | --- | --- | --- | --- |
| Row | Outlier score | Data index | Outlier description | Outlier score | Data index | Outlier description |
| 1 | 0.34153 | 9 | ADMIN/SOCIAL ADMISSION | 14.9298 | 9 | ADMIN/SOCIAL ADMISSION |
|   | 0.23951 | 100 | IMMUNITY DISORDER | 6.4961 | 100 | IMMUNITY DISORDER |
|   | 0.23353 | 105 | INFLUENZA Rank =3 | 6.0001 | 105 | INFLUENZA |
| 5 | 0.33642 | 27 | ADMIN/SOCIAL ADMISSION \| White | 10.6857 | 27 | ADMIN/SOCIAL ADMISSION \| White |
|   | 0.29557 | 562 | OTHR RESPIRATRY CANCER \| Other Race | 5.1053 | 562 | OTHR RESPIRATRY CANCER \| Other Race |
| 6 | 0.33699 | 26 | ADMIN/SOCIAL ADMISSION \| Spanish/Hispanic | 18.1048 | 26 | ADMIN/SOCIAL ADMISSION \| Spanish/Hispanic |
|   | 0.32797 | 690 | SHOCK \| Unknown | 12.2894 | 690 | SHOCK \| Unknown |
| 7 | 0.36111 | 44 | ADMIN/SOCIAL ADMISSION \| 70 or Older | 89.1931 | 44 | ADMIN/SOCIAL ADMISSION \| 70 or Older |
|   | 0.32116 | 361 | GANGRENE \| 18 to 29 | 7.0126 | 361 | GANGRENE \| 18 to 29 |
| 11 | 0.33022 | 721 | GOUT/CRYSTAL ARTHRPTHY \| Unknown \| Black/African American | 5.5963 | 987 | MAL-POSITION/PRESNTATN \| Unknown \| Black/African American |
|   | 0.32183 | 987 | MAL-POSITION/PRESNTATN \| Unknown \| Black/African American | 3.5063 | 75 | ADMIN/SOCIAL ADMISSION \| Spanish/Hispanic \| Other Race |
| 13 | 0.34191 | 116 | ADMIN/SOCIAL ADMISSION \| Not Span/Hispanic \| 70 or Older | 34.2099 | 116 | ADMIN/SOCIAL ADMISSION \| Not Span/Hispanic \| 70 or Older |
|   | 0.33850 | 565 | CHILDHOOD DISORDERS \| Spanish/Hispanic \| 18 to 29 | 6.0413 | 1507 | MEDICAL EXAM/EVALUATN \| Spanish/Hispanic \| 18 to 29 |

Table 8: Comparing the outliers detected by the isolation forest and feature bagging outlier detection techniques. The outlier scores are provided by the individual outlier detection techniques.

D. ***The California OSHPD Dataset***

In Figure 17 we present the results from applying our PIKS technique to the data released by the California OSHPD.

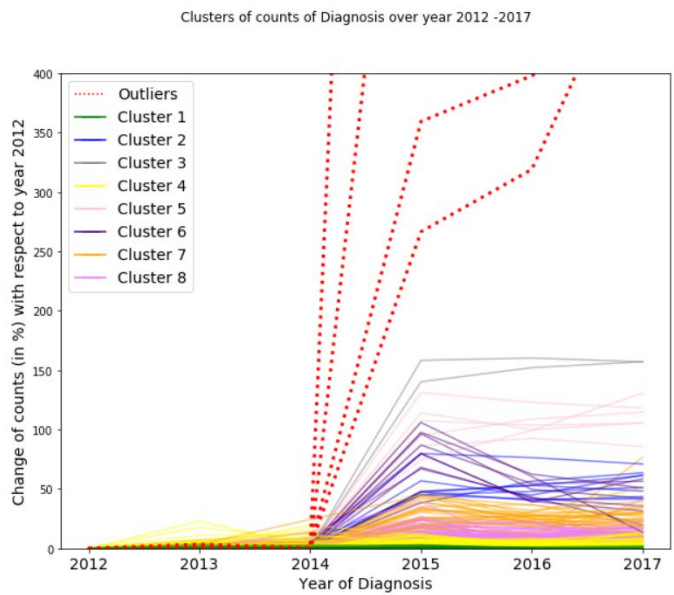

Figure 17: The results of applying the PIKS technique to data released by the California OSHPD from 2012-2018. Outliers are shown in dotted red lines. The other clusters are color coded. The outlier descriptions are provided in

In Figure 18 we present results of using the auto-encoder.

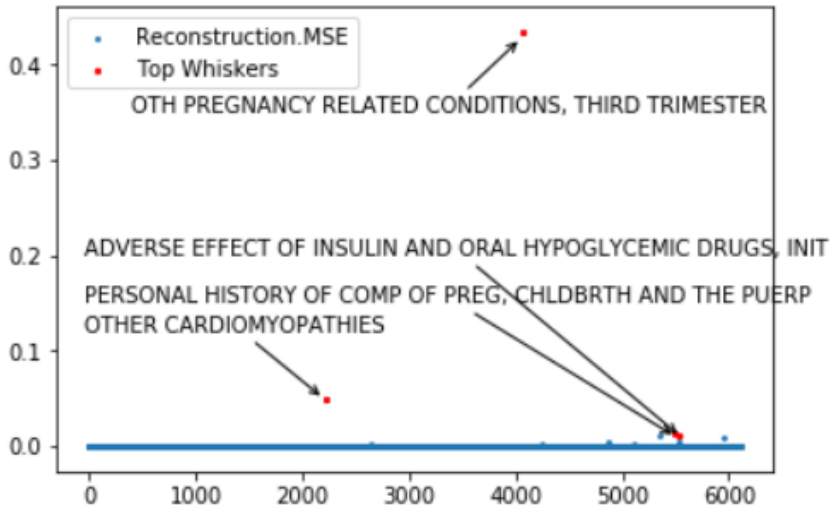

Figure 18: The results of applying the auto-encoder to data released by the California OSHPD from 2012-2017. Outliers are shown in red dots, with the disease descriptions. The x-axis contains the feature index, and the y-axis depicts the reconstruction error. Table 9 compares the outliers detected by the two techniques.

| Outliers detected through the PIKS technique | Outliers detected by the auto-encoder |
|---|---|
| OTHER CARDIOMYOPATHIES | OTH PREGNANCY RELATED CONDITIONS, THIRD TRIMESTER |
| OTH PREGNANCY RELATED CONDITIONS, THIRD TRIMESTER | OTHER CARDIOMYOPATHIES |
| PERSONAL HISTORY OF COMP OF PREG, CHLDBRTH AND THE PUERP | PERSONAL HISTORY OF COMP OF PREG, CHLDBRTH AND THE PUERP |
| ADVERSE EFFECT OF INSULIN AND ORAL HYPOGLYCEMIC DRUGS, INIT | ADVERSE EFFECT OF INSULIN AND ORAL HYPOGLYCEMIC DRUGS, INIT |

Table 9: In this table, the California OSHPD data was analyzed. We compare the top few outliers detected by the subset scan technique with the auto-encoder. The ICD10 descriptors are shown.

## VI. DISCUSSION

From Figure 7, we note that the outliers consist of CCS codes representing "Administrative/Social Admission" and "Influenza". Typically, "Administrative/Social Admissions" are used by emergency departments to classify geriatric populations with unmet needs [65]. Platt-Milles et al. [66] present conclusive analysis about increasing emergency department visits attributed to geriatric patients, and estimate that in North Carolina, such visits will increase 47% from 2007 to 2030. The identification of "Administrative/Social Admissions" as an outlier can lead to a deep dive into the New York SPARCS dataset about specific factors related to this CCS diagnosis code, such as counties and hospitals that experienced this increase. Such detailed analysis is outside the scope of this paper.

From Figure 8, we note that the "Administrative/Social Admission" rates shows a large increase for the race label of "White". Lo et al. [65] have reported race-based differences among emergency room admittances for heart disease. However, the factors that govern this observation are still under investigation. Similarly, Figure 9 shows an increase for the ethnicity of "Spanish/Hispanic". The value of our methodology is that it can be easily re-run with newly released open health data every year, as and when such data is made available by state and federal governments.

From Figure 10, we observe a very large increase in the number of "Administrative/Social Admission" cases for the Age Group 70 and above. The analysis of Platt-Milles et al. [66] was performed for the state of North Carolina, and warned of significant increases in admittance of geriatric populations to emergency departments by year 2030. A similar cautionary tale arises from the plot in Figure 10, and is applicable to the state of New York. We emphasize that this plot was not cherry-picked or hand-crafted from the data, but emerged from automated exploratory analysis carried out by our PIKS analysis.

A recent report by United Health estimates that there are 18 million avoidable hospital emergency visits every year that add $32 billion in healthcare costs per year [67]. The outlier analysis we conducted also shows that these emergency hospital visits are increasing. However, we are not able to determine whether the emergency visit was avoidable, as this level of detail is not provided by the New York SPARCS data. Nevertheless, the plot in Figure 10 conveys the magnitude of this problem to the public in an accessible way.

In Figure 11, we notice a large increase in the CCS code "Mal-position/Presentation", referring to birth problems in Blacks/African American females. Recent research by Pilliod [68] recommends that special training be provided to providers most likely to manage these challenging. Though they did not identify specific geographic locations or hospitals where such a need exists, we can readily obtain this information from the New York SPARCS dataset. Hence, our analysis provides value to policy makers who can decide which efforts require special emphasis, and how to distribute resources geographically.

From Figure 12 we see that "Administrative/Social Admissions" also shows a large increase for Age Group 70 and older who are not Hispanic. This is consistent with our earlier analysis about this CCS diagnosis code. Different ethnic groups are identified through our analysis.

It is interesting that other independent research by Mehrabi et al. [69] has also identified the CCS code for "Administrative/Social Admission" as a code of interest based on a deep learning methodology. The data used was different, and consisted of longitudinal patient records from Olmsted County, Minnesota, with about half a million unique records. Their method produced heatmaps, as shown in Figure 19. Note that in Figure 19, precise changes in variables cannot be observed as clearly as depicted in Figure 7 of our paper. This shows that our technique produces results that are consistent with other research studies while being more interpretable. For instance, we can clearly see in Figure 7 that "Administrative/Social Admission" experienced a 250% increase from 2009 to 2012, and far outpaced increases in all other CCS codes.

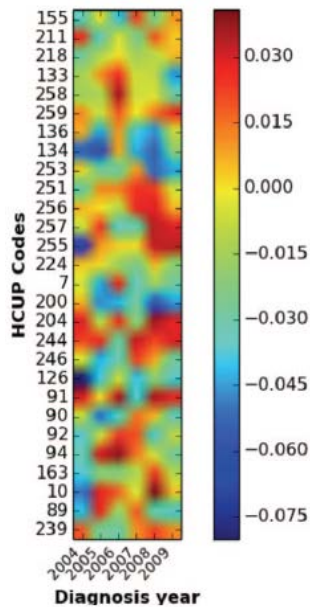

Figure 19: This figure shows the result obtained by Mehrabi et al. [69]. They used a deep neural network to identify temporal patterns in CCS diagnosis codes. The heatmap shown represents the weights of a node in the hidden layer of a 3-layer neural network. Higher weight values are colored in red, and represent connections with CCS codes of interest.

From Figure 14, we can observe that four hospitals have been identified as outliers, as the average costs of different diagnoses are dissimilar compared to the rest of the hospitals. Figure 14 (A) shows that Coler-Goldwater hospital has several CCS diagnoses codes with higher average cost than other hospitals. The Goldwater campus of this hospital was closed in 2013 [70] and the Coler campus was facing imminent closure in 2018 [70]. This shows that trends in the data could be potentially long term, and if not addressed seriously by policy-makers and the local communities, it could have adverse impacts in those regions.

From Figure 14 (B), we note that 'Summit Park Hospital' has a very high average cost of approximately $600,000 for 'personality disorder' treatment. This hospital was closed in 2015 due to cost overruns [71, 72] and 182 people had to be relocated to alternate nursing home facilities. Though the financial situations of hospitals are closely guarded, the public can get a glimpse of their operational status through such open data sets, provided they have the tools to interpret and analyze the data.

From Figure 14 (D) we observe that "North Central Bronx Hospital" has a very high average cost of nearly $700,000 for diagnosis 'Cancer of Bronchus; Lung'. This suggests that specific interventions need to be introduced in this particular community. Indeed, recent efforts have been focused on improving primary care access to the population in this region, specifically in terms of educating the public about lung cancer [73]. Furthermore, hospitals in this region such as St. Barnabas Hospital have recently started early screening procedures for smokers.

From Figure 14 (C), we observe that the 'Bon Secours Community Hospital' has high treatment cost for several diagnoses. This hospital recently underwent consolidation with the Westchester Medical Center [74]. It is likely that the outlier cost distribution in Figure 14 (C), signaled issues at the Bon Secours Health System, which sought resolution through a merger. This is a hypothesis, driven directly by the data and further verification of this line of thinking is required. This is an example of how the data can provide ground truth about the performance of different hospitals, which can be analyzed by the public. This hypothesis testing and verification process is captured at the bottom of Figure 2.

Another hospital identified as an outlier by our analysis is the Henry J. Carter Specialty Hospital in 2011 [75]. This hospital was formed to provide long-term acute care for physically disabled and medically fragile individuals. Given the type of patients that this hospital is designed to serve, it appears reasonable that their cost structure will be different from other hospitals in New York state.

There is a current policy debate around the issue of nonprofit hospitals being more responsive to community health issues. Rozier et al argue [76] for a change in the Internal Revenue Code in the US which will allow hospitals to increase their engagement and collaboration with public health organizations and community-based non-profit organizations. Our analysis has shown that the North Central Bronx region could benefit from such a collaboration. Future analysis of New York SPARCS data will show if such efforts have been able to drive down healthcare costs in this region.

Figure 15 shows outliers in the New Yorks SPARCS data when CCS Diagnosis Codes are aggregated over Age Groups. One of the top outliers consists of "Suicide and intentional self-injury", Age Group 50-69. This observation is similar to the work of Deaton [1], who also showed that the middle-aged population is become increasingly vulnerable. The difference is that Deaton [1] had to manually select the age group of 40-50 years whereas our algorithm automatically identified the Age Group 50-69 as containing an outlier for the CCS Diagnosis Code of "Suicide and intentional self-injury". Though the end results are similar, our approach lends itself to automation and sifting through large datasets efficiently.

We also note that the ranges in the age groups are different, with Deaton [1] identifying the age group 40-50 years. In our case, such an identification is not possible, because the New York SPARCS data does not provide such an age group classification. This is an indication that in the future, the New York SPARCS organization can work towards releasing data with a finer level of discretization. This is an example of how our analysis can help policy-makers decide what type of data to make available to the public.

Next, we interpret the importance of the outliers we have detected. One of the top outliers in Table 7 is suicides. Recent research has shown a sharp increase in mental health issues amongst adolescents [77, 78], and this is reflected in higher suicide rates. Twenge et al. [77] mention the increasing use of social media as a potential cause. The work of Case and Deaton [1] showed that increased alcohol and opioid usage contributed to increased mortality rates for middle aged white non-Hispanic Americans. Though the data we analyzed was specifically for the state of New York, the trends we detected are consistent with analysis of national US data.

Another top outlier in Table 7 is immunity disorders. Researchers such as Irwin and Miller [79] have shown that mental health disorders such as depressive disorders are linked to reduced immunity. Though the work of Irwin and Miller was done in 2007, our results produced with data in the 2009-2016 timeframe indicate that these two factors are still linked. This suggests that as a public policy issue, more attention needs to be paid to the root causes of mental disorders in society, and how to best help vulnerable populations.

Note that one of the top two outliers in Table 9 is "OTH PREGNANCY RELATED CONDITIONS, THIRD TRIMESTER". This is a significant finding, as earlier research has shown that "maternal mortality rates rose markedly from 2002 to 2006 in California" [80]. According to the authors of that study [80], this prompted an in-depth review of the maternal mortality in the state California. Our results indicate that this continues to remain a problem even in the period 2012-2017 that we analyzed. This suggests that sustained efforts need to be maintained to take care of this health condition. Some of the contributing factors identified in the study [80] included race, substance abuse and obesity. This provides guidance to policy makers about how special emphasis needs to be placed on handling vulnerable populations. For instance, Moussa et al. [81] note that the effects of obesity are far reaching, and also affect maternal mortality rates. One suggested step is to provide preconception counseling to achieve weight loss before pregnancy [81].

Our comparison study summarized in Table 7 and Table 9 indicates that the PIKS outlier detection technique and the auto-encoder produce similar results. This consistency is encouraging. The top three descriptors for the New York SPARCS dataset are the same, as shown in Table 7. Similarly, for the California OSHPD dataset, the top four ICD10 descriptors are the same, as shown in Table 9.

In Table 6, the outliers detected at the first aggregation level are nearly the same for the PIKS technique and the auto-encoder. For multiple aggregation levels, there is a difference, which arises due to the tuning of the auto-encoder. We can produce fewer outliers by increasing the sensitivity of the auto-encoder, as shown in Figure 16. The investigation of the tuning of the auto-encoder is outside the scope of this paper. If policy agencies, hospitals or insurance companies are interested in using this technique, it is likely that they would conduct a deeper analysis of this tuning to suit their business needs. The purpose of the current paper is to demonstrate the utility of automatic outlier detection for conducting exploratory data analysis.

Given that the results between the two techniques are consistent, we briefly examine the operation of the techniques. For the searchlight subset-scan technique, the only parameter that needs to be selected by the user is k, the expected number of items in a cluster. Note that the heart of the searchlight subset-scan technique is an iterative k-means clustering algorithm. We used a value of k=8. We tried values of k in the range 6-8 and found the results did not vary significantly. In contrast, the auto-encoder has several parameters that need tuning. This includes the number of hidden layers, the size of each hidden layer, the size of the epoch and the optimization algorithm used to train the network. It may take considerable time to select the best parameters, and this is also dependent on the type of data analyzed. Hence, the searchlight subset-scan technique is advantageous as such a detailed parameter selection is not required.

Another advantage of our PIKS outlier detection technique is that no training or model-fitting is required, as is the case for the auto-encoder.

An interesting outcome of the application of our outlier detection techniques to open health data is that the broad research findings such as in [80] and [81] can be "fact checked" at a high level. The detailed statistical analysis presented in these papers [80] [81] requires access to a much more fine grained level of data, which are unlikely to be made publicly accessible due to privacy and regulatory concerns. Hence, having an independent and open data source, even though it may be coarse, is invaluable to spreading public awareness of the specific health issues identified by medical researchers. This is especially true in the world today, where the public is becoming increasingly skeptical about health findings and medical research due to many conflicts of interest that have been recently reported. For instance, the chief medical officer of the prestigious Memorial Sloan Kettering Cancer Center in New York resigned after it was found that competing interests were not properly disclosed to the public [82]. Ultimately, as noted by Baack [22], the use of open data and its analysis by concerned citizens is important to allow democracies to function efficiently, and to allow people to arrive at their own interpretation of data surrounding public issues.

Our results demonstrate that interesting and relevant trends in large healthcare related datasets can be efficiently identified using artificial intelligence and machine learning techniques. This provides a valuable methodology and tool for concerned citizens to interpret developments in the field of healthcare and other areas where open data is becoming increasingly available. Such an unbiased data-driven interpretation of breaking news events will become increasingly important in the future. Artificial intelligence and machine learning technologies will play a key role in this process.

The use of open source tools is increasing in the fields of medicine and healthcare. Sonntag has used a collection of open source tools for performing text analysis of unstructured medical data [83]. In order to aid the dissemination of this work, we are making our code freely available at *www.github.com/fdudatamining/*.

VII. CONCLUSION

We presented an efficient outlier detection technique called searchlight subset-scan and applied it to large open health datasets. Although many government agencies are increasingly releasing open health data, there are few open toolkits devoted to exploring and analyzing this data. Without the availability of these toolkits, it becomes difficult for researchers and concerned citizenry to extract the full value from the data. Since exploratory data analysis is an important step, we applied our searchlight subset-scan (PIKS) technique to data released by New York Statewide Planning and Research Cooperative System (SPARCS), and a dataset obtained from California's Office of Statewide Health Planning and Development. The SPARCS data contains 19 million de-identified patient records released by New York State SPARCS from the years 2009 to 2016.

Our technique identified interesting and meaningful trends in the occurrence of different diagnoses including suicide rates, immunity disorders and cardiomyopathies. We implemented and applied three other outlier detection techniques, consisting of auto-encoders, isolation forests and feature bagging. A comparison of the PIKS technique against auto-encoders, isolation forests and feature bagging showed that the top few outliers are similar. There are far fewer parameters to tune for our PIKS technique than the auto-encoder. Furthermore, the PIKS technique requires no training or model fitting as compared with the auto-encoder. Hence, the PIKS technique is advantageous for fast, "out-of-the-box" data exploration.

This outliers identified by our technique can be further investigated by policy makers for targeted public health interventions. Since our approach is scalable and readily replicable, it should be valuable to researchers, and concerned citizens interested in utilizing open health data. The use of artificial intelligence and machine learning techniques greatly expands our ability to deal with large datasets and extract meaningful insights.

ACKNOWLEDGMENT

This research did not receive any grant from funding agencies in the public, commercial, or non-profit sectors.

CONFLICT OF INTEREST

On behalf of all authors, the corresponding author states that there is no conflict of interest.